%% file: master.tex
\documentclass[11pt,a4paper]{article}
\usepackage[hyperref]{acl2018}
\usepackage{times}
\usepackage{latexsym}
\usepackage{graphicx}
\usepackage{url}
\usepackage{multirow}
\usepackage{amsmath}

\usepackage[normalem]{ulem}
\definecolor{darkgreen}{rgb}{0.0, 0.5, 0.0}

\def\MDdel#1{\bgroup\markoverwith{\textcolor{darkgreen}{\rule[0.5ex]{2pt}{1pt}}}\ULon{#1}}

\def\SAdel#1{\bgroup\markoverwith{\textcolor{red}{\rule[0.5ex]{2pt}{1pt}}}\ULon{#1}}

\def\EGdel#1{\bgroup\markoverwith{\textcolor{blue}{\rule[0.5ex]{2pt}{1pt}}}\ULon{#1}}

\aclfinalcopy 

\title{Char2char Generation with Reranking for the E2E NLG Challenge}

\author{
\resizebox{\textwidth}{!}{
\begin{tabular}{c}
Shubham Agarwal \\
 {\normalfont Heriot-Watt University \thanks{\ \ Work done during internship at Naver Labs (Previously \textit{Xerox Research Centre Europe}.)}} \\
 {\normalfont   Edinburgh, UK } \\
  {\tt sa201@hw.ac.uk}
  \end{tabular} 
  \begin{tabular}{c}
  Marc Dymetman \\
 {\normalfont   NAVER Labs Europe\thanks{\ \ Previously \textit{Xerox Research Centre Europe}.}  }\\
 {\normalfont   Grenoble, France }\\
  {\tt marc.dymetman@naverlabs.com}
   \end{tabular}  
  \begin{tabular}{c}
  \'{E}ric Gaussier \\
 {\normalfont  Universit\'{e} Grenoble Alpes} \\
 {\normalfont  Grenoble, France} \\
  {\tt Eric.Gaussier@imag.fr }
   \end{tabular}}
  }

\date{}

\begin{document}
\maketitle
\input abstract.tex
\input introduction.tex
\input model.tex

\input experiments.tex

\input evaluation.tex
\input analysis.tex
\input conclusion.tex
\section*{Acknowledgements}
We like to thank Chunyang Xiao and Matthias Gall\'{e} for their useful suggestions and comments. 

\bibliography{challenge_2017,challenge_2017_2}
\bibliographystyle{acl_natbib}

\end{document}

%% file: abstract.tex
\begin{abstract}

This paper describes our submission to the E2E NLG Challenge. Recently, neural seq2seq approaches have become mainstream in NLG, often resorting to pre- (respectively post-) processing \textit{delexicalization} (relexicalization) steps at the word-level to handle rare words. By contrast, we train a simple character level seq2seq model, which requires no pre/post-processing (delexicalization, tokenization or even lowercasing), with surprisingly good results. For further improvement, we explore two re-ranking approaches for scoring candidates. We also introduce a synthetic dataset creation procedure, which opens up a new way of creating artificial datasets for Natural Language Generation.

\end{abstract}

%% file: introduction.tex
\section{Introduction}

Natural Language Generation from Dialogue Acts involves generating human understandable utterances from slot-value pairs in a Meaning Representation (MR). This is a component in Spoken Dialogue Systems, where recent advances in Deep Learning are stimulating interest towards using end-to-end models. Traditionally, the Natural Language Generation (NLG) component in Spoken Dialogue Systems has been rule-based, involving a two stage pipeline: `sentence planning' (deciding the overall structure of the sentence) and `surface realization' (which renders actual utterances using this structure). The resulting utterances using these rule-based systems tend to be rigid, repetitive and limited in scope. Recent approaches in dialogue generation tend to directly learn the utterances from data \citep{mei2015talk,lampouras2016imitation,duvsek2016sequence,wen2015semantically}.

Recurrent Neural Networks with gated cell variants such as LSTMs and GRUs \citep{hochreiter1997long,cho2014learning} are now extensively used to model sequential data. This class of neural networks when integrated in a Sequence to Sequence \citep{cho2014learning,sutskever2014sequence} framework have produced state-of-art results in Machine Translation \citep{cho2014learning,sutskever2014sequence,bahdanau2014neural}, Conversational Modeling \citep{vinyals2015neural}, Semantic Parsing \citep{xiao2016sequence} and Natural Language Generation \citep{wen2015semantically,mei2015talk}. While these models were initially developed to be used at word level in NLP related tasks, there has been a recent interest to use character level sequences, as in Machine Translation \citep{chung2016character,zhao2016efficient,ling2015character}. 

Neural seq2seq approaches to Natural Language Generation (NLG) are typically word-based, and resort to delexicalization (a process in which named entities (slot values) are replaced with special `placeholders' \citep{wen2015semantically}) to handle rare or unknown words (out-of-vocabulary (OOV) words, even with a large vocabulary). It can be argued that this de-lexicalization is unable to account for phenomena such as morphological agreement (gender, numbers) in the generated text \citep{sharma2016natural,nayak2017}. 

However, \newcite{Goyal2016} and \newcite{Agarwal2017} employ a char-based seq2seq model where the input MR is simply represented as a character sequence, and the output is also generated char-by-char; avoiding the rare word problem, as the character vocabulary is very small. This work builds on top of the formulation of \newcite{Agarwal2017} and describes our submission for the E2E NLG challenge \citep{novikova2017e2e}. We further explore re-ranking techniques in order to identify the perfect `oracle prediction' utterance. One of the strategies for re-ranking uses an approach similar to the `inverted generation' technique of \citep{chisholm2017learning}. \newcite{sennrich2015improving}, \newcite{li2015diversity} and \newcite{konstas2017neural} have also trained a reverse model for back translation in Machine Translation and NLG. A synthetic data creation technique is used by \newcite{duvsek2017referenceless} and \newcite{logacheva2015role} but as far as we know, our protocol is novel. Our contributions in this paper and challenge can, thus, be summarized as: 
\vspace{-2mm}
\begin{enumerate}
\item We show how a vanilla character-based sequence-to-sequence model performs successfully on the challenge test dataset in terms of BLEU score, while having a tendency to omit semantic material. As far as we know, we are the only team using character based seq2seq for the challenge.
\vspace{-2mm}
\item We propose a novel data augmentation technique in Natural Language Generation (NLG) which consists of `editing' the Meaning Representation (MR) and using the original ReFerences (RF). This fabricated dataset helps us in extracting features (to detect errors), used for re-ranking the generated candidates (\mbox{Section \ref{sub-sect:dataAugment}}).
\vspace{-2mm}
\item We introduce two different re-ranking strategies corresponding to our primary and secondary submission (in the challenge), defined in \mbox{Section \ref{sub-sect:reranking}}.\footnote{Due to space limitations, our description here omits a number of aspects. 
For a more extensive description, analysis and examples, please refer to \url{http://www.macs.hw.ac.uk/InteractionLab/E2E/final_papers/E2E-NLE.pdf}.}
\end{enumerate}

%% file: model.tex
\section{Model}
\label{sect:Model}
\vspace{-1mm}
In the sequel, we will refer to our vanilla char2char model with the term Forward Model. 
\vspace{-1mm}
\subsection{Forward Model}

We use a Character-based Sequence-to-Sequence RNN model \citep{sutskever2014sequence,cho2014learning} with attention mechanism \citep{bahdanau2014neural}. We feed a sequence of embeddings of the individual characters composing the source Meaning Representation (MR) -seen as a string- to the Encoder RNN and try to predict the character sequence of the corresponding utterances (RF) in the generation stage with the Decoder RNN.  

Coupled with the attention mechanism, seq2seq models have become de-facto standard in generation tasks. The encoder RNN embeds each of the source characters into vectors exploiting the hidden states computed by the RNN. The decoder RNN predicts the next character based on its current hidden state, previous character, and also the ``context'' vector $c_i$, computed by the attention model. 


While several strategies have been proposed to improve results using Beam Search in Machine Translation \citep{freitag2017beam}, we used the length normalization (aka length penalty) approach \newcite{wu2016google} for our task. A heuristically derived length penalty term is added to the scoring function which ranks the probable candidates used to generate the best prediction. 

%
%

\subsection{Protocol for synthetic dataset creation}
\label{sub-sect:dataAugment}
We artificially create a training set for the classifier (defined in Section \ref{sect-classifier}) to detect errors (primarily omission of content)  in generated utterances, by a data augmentation technique. The systematic structure of the slots in MR gives us freedom to naturally augment data for our use case. To the best of our knowledge, this is the first approach of using data augmentation in the proposed fashion and opens up new directions to create artificial datasets for NLG. We first define the procedure for creating a dataset to detect omission and then show how a similar approach can be used to create a synthetic dataset to detect additions.

\textit{Detecting omissions.} This approach assumes that originally there are no omissions in RF for a given MR (in the training dataset). These can be considered as positive pairs when detecting omissions. Now if we artificially add another slot to the original MR and use the same RF for this new (constructed) MR, naturally the original RF tends to show omission of this added slot. 
 \vspace{-0.5mm}
 \begin{equation}
\begin{split}
MR_{original} \xrightarrow{\text{+ Added slot}} MR_{new} \\
%
MR_{original} \xrightarrow{\text{- Removed slot}} MR_{new}
\end{split}
\end{equation}
\vspace{-0.5mm}
This is a two stage procedure: (a) Select a slot to add. (b) Select a corresponding slot value. Instead of sampling a particular slot in step (a), we add all the slots one by one (that could be augmented in MR apart from currently present slots). Having chosen the slot type to be added, we add the slot value according to probability distribution of the slot values for that slot type. The original (MR$_{original}$,RF) pair is assigned a class label of 1 and the new artificial pairs (MR$_{new}$,RF) a label of 0, denoting a case of omission (first line of (1)). Thus, these triplets (MR, RF, Class Label) allow us to treat this as a classification task. 


\textit{Detecting additions.} In order to create a dataset which can be used for training our model to detect additions, we proceed in a similar way. The difference is that now we systematically remove one slot in the original MR to create  the new MRs (second line of (1)). 

In both cases, we control the procedure by manipulating MRs instead of the Natural Language RF. This kind of augmented dataset opens up the possibility of using any classifier to detect the above mentioned errors. 

\subsection{Re-ranking Models}
\label{sub-sect:reranking}

In this section, we define two techniques to re-rank the n-best list and these serve as primary and secondary  submissions to the challenge. 

\subsubsection{Reverse Model}

\begin{figure*}[ht]
\centering{\includegraphics[scale=.5]{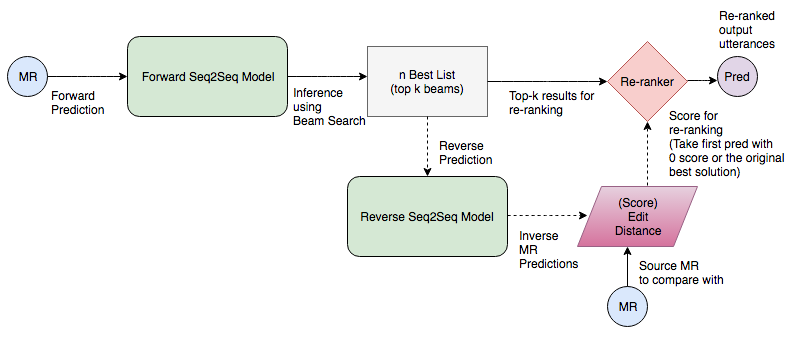}}
\caption{Illustration of the pipeline for the re-ranking approach (based on inverse reconstructions using reverse model) as described in Section \ref{sub-sect:reranking}. Apart from Forward and Reverse seq2seq models, we have a re-ranker based on the edit distance of the actual MR and the inverse reconstructed MR.}
\label{fig:ReversePipeline}
\end{figure*}

We generated a list of top-k predictions (using Beam Search) for each MR in what we call the \textit{forward} phase of the model. In parallel, we trained a \textit{reverse} model which tries to reconstruct the MR given the target RF, similar to the autoencoder model by \newcite{chisholm2017learning}. This is guided by an intuition that if our prediction omits some information, the reverse reconstruction of MR would also tend to omit slot-value pairs for the omitted slot values in the prediction. We then score and re-rank the top-k predictions based on a distance metric, namely the edit distance between the original MR and the MR generated by the reverse model, starting from the utterance predicted in the forward direction. 

To avoid defining the weights when combining edit distance with the log probability of the model, we used a simplified mechanism. At the time of re-ranking, we choose the first output in our n-best list with zero edit distance as our prediction. If no such prediction can be found, we rely upon the first prediction in our (probabilistically) sorted n-best list. Figure \ref{fig:ReversePipeline} illustrates our pipeline approach.

\subsubsection{Classifier as a re-ranker}
\label{sect-classifier}
To treat omission (or more generally any kind of \textit{semantic adequacy} mis-representation such as repetition or addition of content) in the predictions as a classification task, we developed a dataset (consisting of triplets) using the protocol defined earlier. However, to train the classifier we relied on hand-crafted features based on string matching in the prediction (with corresponding slot value in the MR). In total, there were 7 features, corresponding to each slot (except `name' slot).
To maintain the class balance, we replicated the original (MR,RF) pair (with a class label of 1) for each artificially generated (MR,RF) pair (with a class label of 0, corresponding to omissions). 


We used a logistic regression classifier to detect omissions following a similar re-ranking strategy as defined for the reverse model. For each probable candidate by the forward model, we first extracted these features and predicted the label by this logistic regression classifier. The first output in our n-best list with a class label 1 is then chosen as the resulting utterance. As a fallback mechanism, we rely on the best prediction by the forward model (similar to the reverse model). We chose the primary submission to the challenge as the pipeline model with classifier as re-ranker. Our second submission was based on re-ranking using the reverse model while the vanilla forward char2char model was our third submission.

%% file: experiments.tex
\section{Experiments}


The updated challenge dataset comprises 50K canonically ordered and systematically structured (MR,RF) pairs, collected following the crowdsourcing protocol explained in \newcite{novikova2016crowd}. 
Consisting of 8 different slots (and their respective different values), note that the statistics in the test set differ significantly from the training set. We used the open source \textit{tf-seq2seq} framework\footnote{\url{https://github.com/google/seq2seq}\ .}, built over TensorFlow \citep{abadi2016tensorflow} and provided along with \citep{britz2017massive}, with some standard configurations.
We experimented with different numbers of layers in the encoder and decoder as well as different beam widths, while using the bi-directional encoder with an ``additive'' attention mechanism. In terms of BLEU, our best performing model had the following configuration: encoder 1 layer, decoder 2 layers, GRU cell, beam-width 20, length penalty 1. 

%% file: evaluation.tex
\section{Evaluation}

We chose our primary system to be the re-ranker using the classifier. Table \ref{table:test-bleu} summarizes our ranking among all the 60+ submissions (primary as well as additional) on the test set. In terms of BLEU, two of our systems were in the top 5 among all 60+ submissions to the challenge.


\begin{table}[h]
\centering
\resizebox{0.48\textwidth}{!}{
\begin{tabular}{llc }
  \hline
Submission & BLEU & Overall Rank \\
  \hline \hline
Re-ranking using classifier (Primary) & 0.653 & 18 \\
Re-ranking using reverse (Secondary) & 0.666 & 5 \\
Forward (Third) & 0.667 & 4 \\
Baseline & 0.659 & 10 \\
 \hline
\end{tabular}}
\caption{Automatic BLEU evaluations released by organizers on the final challenge submission. We had 3 submissions as described in Section \ref{sect:Model}. Two of our systems were in the top 5 among all 60+ submissions.}
\label{table:test-bleu}
\end{table}



\begin{table}[h]
\centering
\resizebox{0.38\textwidth}{!}{
\begin{tabular}{ccccc}
  \hline
Metric & TrueSkill & Range & Cluster \\
  \hline \hline
Quality & 0.048 & (8-12) & 2 \\
Naturalness & 0.105 & (4-8) & 2 \\
 \hline
\end{tabular}}
\caption{Human evaluation was crowd-sourced on the primary system according to the TrueSkill algorithm \citep{Sakaguchi2014}}
\label{table:test-human}
\end{table}
		
Results for human evaluation, as released by the challenge organizers, are summarized in Table \ref{table:test-human}. They followed the TrueSkill algorithm \citep{Sakaguchi2014} judging all the primary systems on \textit{Quality} and \textit{Naturalness}. We obtained competitive results in terms of both metrics, our system being in the 2nd cluster out of 5 (for both evaluations). On the other hand, most systems ranked high on quality tended to have lower ranks for naturalness and vice versa.

%% file: analysis.tex
\section{Analysis}

We found that the presence of an `oracle prediction' (perfect utterance) was dependent on the number of slots in the MR. When the number of slots was 7 or 8, the presence of an oracle in the top-20 predictions decreased significantly, as opposed to the case when the number of slots was less than 7. However, the most prominent issue was that of omissions, among the utterances produced in first position (by forward model). There were no additions or non-words. We observed a similar issue of omissions in human references (target for our model) as well. Our two different strategies, thus, improved the semantic adequacy by re-ranking the probable candidates and successfully finding the `oracle' prediction in the top-20 list. However, in terms of automatic evaluation, the BLEU score showed an inverse relationship with adequacy. Nevertheless, we chose our primary system to be the re-ranker with a classifier over the forward model. 

We did not find any issues while ``copying'' the restaurant `name' or `near' slots on the dev set. However, on the test set, as the statistics of the data changed in terms of both slots, we found a tendency of the model to generate the more frequent slot values (corresponding to both slots in the training dataset), instead of copying the actual slot value. 

%% file: conclusion.tex
\section{Conclusion}

We show how a char2char model can be employed for the task of NLG and show competitive results in this challenge. Our vanilla character based model, building on \newcite{Agarwal2017}, requires minimal effort in terms of any processing of dataset while also producing great diversity in the generated utterances. We then propose two re-ranking strategies for further improvements. Even though re-ranking methods show improvements in terms of semantic adequacy, we find a reversal of trend in terms of BLEU.

Our synthetic data creation technique could be adapted for augmenting NLG datasets and the classifier-based score could also be used as a reward in a Reinforcement Learning paradigm.